\documentclass{article}

\usepackage{microtype}
\usepackage{graphicx}
\usepackage{subfigure}
\usepackage{booktabs} 

\usepackage[hyphens]{url}
\usepackage{hyperref}
\usepackage{cleveref}
\usepackage{textcomp}

\usepackage[accepted]{icml2020}

\icmltitlerunning{Monitoring and explainability of models in production}

\begin{document}

\twocolumn[
\icmltitle{Monitoring and explainability of models in production}



\icmlsetsymbol{equal}{*}

\begin{icmlauthorlist}
\icmlauthor{Janis Klaise}{equal,seldon}
\icmlauthor{Arnaud Van Looveren}{equal,seldon}
\icmlauthor{Clive Cox}{seldon}
\icmlauthor{Giovanni Vacanti}{seldon}
\icmlauthor{Alexandru Coca}{seldon}
\end{icmlauthorlist}

\icmlaffiliation{seldon}{Seldon Technologies Ltd, London, United Kingdom}

\icmlcorrespondingauthor{Janis Klaise}{jk@seldon.io}
\icmlcorrespondingauthor{Arnaud Van Looveren}{avl@seldon.io}

\icmlkeywords{Machine Learning, ICML}

\vskip 0.3in
]



\printAffiliationsAndNotice{\icmlEqualContribution} 

\begin{abstract}
The machine learning lifecycle extends beyond the deployment stage. Monitoring deployed models is crucial for continued provision of high quality machine learning enabled services. Key areas include model performance and data monitoring, detecting outliers and data drift using statistical techniques, and providing explanations of historic predictions. We discuss the challenges to successful implementation of solutions in each of these areas with some recent examples of production ready solutions using open source tools.
\end{abstract}

\section{Introduction}\label{sec:intro}
In recent years, the problem of how to deploy and scale machine learning models has been made easier by open source tools such as Seldon Core \cite{seldon-core}, KFServing \cite{kfserving}, Kubeflow \cite{kubeflow} and MLFlow \cite{mlflow}. One of the big challenges in MLOps is to design systems that monitor live deployments and take action or raise alerts when events impacting model performance are encountered \cite{diethe2019continual}. Traditional application monitoring involves logging such core metrics as request latency, frequency and server load with the intention of ensuring uninterrupted, high quality service. On the other hand, it is more demanding to maintain the same level of service for machine learning applications. We identify four additional areas following model deployment which are key for the success of the application:

\begin{enumerate}
    \item Monitoring model performance
    \item Monitoring metrics related to incoming data
    \item Detecting outliers and drift
    \item Explaining model predictions
\end{enumerate}

Firstly, it is critical to ensure model performance does not degrade in a production setting. Inability to detect model performance degradation can lead to stale models and increased technical debt \cite{breck2017, sculley2015}. Whilst trained models usually come with performance metrics on offline test sets, this does not guarantee similar performance in live systems.

Secondly, measuring model performance implies having timely access to labels for live data which are seldom available due to operational and financial constraints. In the absence of labels it is critical to monitor the statistics of input data and output predictions as these can serve as a proxy for model performance \cite{breck2017}.

Thirdly, to be truly useful the monitoring system requires functionality to determine when significant changes to data and predictive distributions happen, also known as drift detection. A related task is to identify incoming data points which fall outside the training data distribution, also known as outlier detection. These questions are statistical in nature and often require separate models which makes it more difficult to provide general solutions.

Finally, it is important to build trust in machine learning systems and make the decision process transparent. Many models are often ``black boxes" whose internal processing is not well understood even by trained data scientists. The field of explainable AI provides numerous approaches to the problem of explaining model predictions, some of which are uniquely suited for deployed models.

In the following sections we discuss the key challenges in these areas from an algorithmic as well as infrastructure perspective and point to existing solutions using open source technologies. 

\section{Monitoring}\label{sec:monitoring}

\subsection{Performance and metrics}\label{sec:metrics}
Ensuring high model performance in live deployments is arguably the most important aspect of monitoring machine learning systems. However, performance metrics used to develop the models depend on the availability of labels. For live data the number and frequency of collected labels depends on the application. For example, in high-frequency time-series prediction or internet ad serving labels are automatic and near real-time whilst for many other applications labels are expensive to produce, potentially unreliable and with long time delays following a prediction. For instance, consider a medical diagnostic system based on image recognition. Labelling even small samples of the data presented to such a system can be challenging since it is time consuming and requires domain knowledge. Label noise can also occur due to disagreements between annotators \cite{louie2010inter, khoo2012prostate, bridge2016intraobserver}. It is however still crucial to report metrics which inform whether the performance of the deployed model is satisfactory.

In the perfect scenario of full and immediate label availability the calculation of performance metrics is straightforward, however some challenges remain on the infrastructure and decision making level. Firstly, the labels have to be fed into some system that then calculates metrics on the fly. This can be a separate system or as part of the model deployment. For example, Seldon Core provides a dedicated \texttt{/send-feedback} API endpoint accepting labels and performing user-defined metric calculations which can be standard ML metrics (e.g. accuracy for classification models) or custom, business-specific metrics (e.g. key KPIs derived from performance metrics). Secondly, metrics are inherently stateful---updating a metric with new values involves using the metric value from the previous step. This is in contrast with deployed models which are static over the lifetime of the model. Whilst recent open source libraries provide solutions to such \emph{online} learning of metrics and histograms \cite{creme}, the engineering effort of ensuring a metrics component does not lose its state and does not go out of sync (e.g. if the metrics component is scaled up to multiple instances) remains. Thirdly, one must decide on appropriate time periods for calculating metrics. For some use cases metrics over the lifetime of the deployment are most useful while for others static or dynamic time windows based on real time or request frequency are more appropriate. Finally, even with accurate knowledge of label-dependent metrics one must set some threshold or decision rule for when to raise alerts when metrics deteriorate. Such thresholds require domain knowledge and can be difficult to set appropriately to limit the number of false alarms. Alternatives based on change-point detection may prove to be more robust \cite{bifet2017classifier}.

In the more common scenario of scarce labels it is common to use label-independent metrics as a proxy for model performance \cite{breck2017}. This includes designing metrics for monitoring live input data and model predictions. In contrast to performance metrics, appropriate metrics depend on the data type. Image, text and tabular data will have different metrics of interest. Even in the widespread case of tabular data, metrics will be different based on the type of features. For example, whilst we can monitor the feature-wise moments and order statistics of numerical features, frequency based measures are more appropriate for categorical features. Going beyond pointwise metrics, fast algorithms for approximate online histograms \cite{histograms2010} can provide much richer insight into the live data distributions. Nevertheless, a fundamental drawback of univariate metrics is that correlations between features are not captured. Multivariate metrics such as covariance matrices and multivariate histograms remain difficult to implement due to the increased computational cost and curse of dimensionality as well as the need for an online update rule. We discuss multivariate approaches in the context of detecting drift in \Cref{sec:cd}.

\subsection{Outlier detection}\label{sec:od}

Machine learning models often fail to generalize outside of the training data distribution \cite{recht2019, engstrom, hendrycks}. Furthermore, models are typically not well calibrated \cite{guocalibration} which can lead to overconfident predictions on out-of-distribution instances \cite{Lakshminarayanan}. Outlier detection is therefore key to flag anomalies whose model predictions we cannot trust and should not use in a production setting. The type of outlier detector for a specific application depends on the modality and dimensionality of the data, availability of labeled normal and outlier data, and whether the detector is pre-trained (offline) or updated online. The pre-trained detector can be deployed as a separate static machine learning model while the online detector is deployed as a stateful application. Labeled outlier data is often scarce, making the problem semi-supervised at best. Open source libraries such as Alibi Detect \cite{alibi-detect} and PyOD \cite{zhao2019pyod} provide a wide range of mainly unsupervised off-the-shelf outlier detectors which can be tailored to the specific problem setting.

It is important to note that the problem of unsupervised anomaly detection for real-world data (e.g. natural images or noisy time series) is far from solved. At the heart the problem is one of reliable density estimation. The quality of the estimator however depends on the modality and the data set. Different studies on image data \cite{nalisnick2018do, choi2018waic} also illustrate that generative density models can assign higher likelihood values to out-of-distribution instances compared to inlier data. Contrary to most machine learning tasks, the unsupervised training objective is only a proxy for the actual performance of the detector.




\subsection{Drift detection}\label{sec:cd}


While outliers refer to individual instances, data drift or shift detection checks whether two samples are drawn from the same underlying distribution or not via a statistical hypothesis test. The goal of the drift detector is therefore to identify when the distribution of the requests for the deployed model starts to diverge from the training data and model predictions become unreliable. We can further distinguish \emph{covariate shift} from \emph{label shift} of the model predictions. In the case of covariate shift the input data distribution $p(x)$ changes while the conditional label distribution $p(y|x)$ remains unchanged. Label shift happens when $p(y)$ changes but the conditional $p(x|y)$ does not.

In order to make drift detection work in practice on high-dimensional data such as images, the incoming data first undergoes a dimensionality reduction step before applying the hypothesis test. \citet{rabanser2019} observe that randomly initialized encoders and \emph{black box dimensionality reduction} introduced by \citet{lipton2018detecting} are promising pre-processing methods. This is followed by a two-sample test such as the \emph{maximum mean discrepancy} \cite{grettonmmd} for the multivariate case in combination with a permutation test to obtain p-values. Alternatively, a feature-wise Kolmogorov-Smirnov test \cite{kstest} is run with Bonferroni or False Discovery Rate \cite{fdr} p-value correction for multivariate data.

Drift detection informs the user when the model should be retrained which is especially important in applications where model performance feedback is not readily available. Malicious data drift, which significantly harms model performance, is of special importance. \citet{vacanti} show that drift detection on an instance level adversarial score is effective at identifying underlying malicious data shift. Alibi Detect \cite{alibi-detect} provides all the aforementioned drift detection functionality under a common API.

\subsection{Deploying model monitoring}

\begin{figure}[htbp]
    \centering
    \includegraphics[width=\columnwidth]{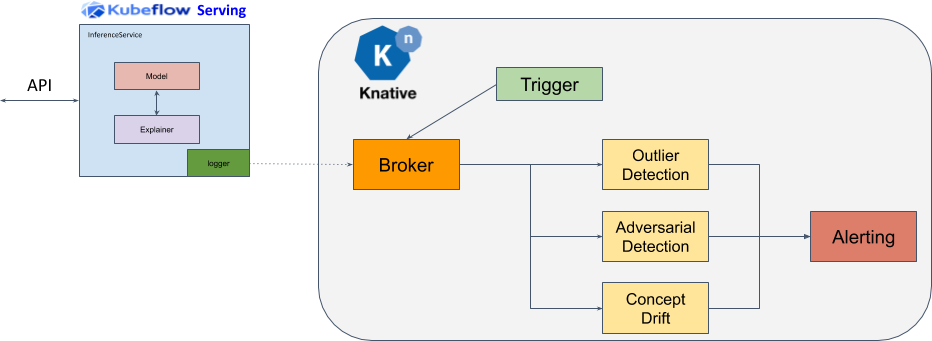}
    \caption{Asynchronous ML monitoring implementation with KFServing and KNative eventing.}
    \label{fig:kfserving_monitoring}
\end{figure}

The techniques discussed above need to be deployed alongside the running models but in a manner which does not adversely affect their core performance. Recent inference deployment projects such as KFServing and Seldon Core which run on the Kubernetes container orchestration platform solve this by utilizing the eventing based project KNative \cite{knative} which allows serverless components to be connected to event streams. The serverless KNative project makes it possible to scale services down to zero if no requests are being received and scale up as demand increases. This allows monitoring components to scale as needed. \Cref{fig:kfserving_monitoring} shows the resulting architecture.

Incoming low latency requests run as normal with a payload logging solution sending events containing model request and response payloads to a KNative \emph{broker} which can farm these out as desired via programmable \emph{triggers} to serverless components such as outlier, drift and adversarial detection modules. Further eventing components can be added to feed off events produced by these components to send onwards to, for example, alerting or storage modules. The architecture provides a clean separation of concerns between the model and its later analysis components each of which can be scaled separately.

\section{Explainability}\label{sec:exp}

\subsection{The need for model explanations}\label{sec:bg}

Instance level machine learning model explanations are desirable for multiple reasons. They allow the user to build trust in the predictions made by the deployed machine learning system and improve transparency. The user can verify which factors contributed to certain predictions, introducing a layer of accountability. Model introspection is also increasingly important with the rise of pre-trained models (e.g. \citet{alex2019big} for computer vision or \citet{Wolf2019HuggingFacesTS} for language tasks) which are only fine-tuned for a specific downstream task. The data used during the pre-training stage is not curated by the user of the downstream task and can create harmful model biases, potentially leading to unfair outcomes. \citet{joybuolamwini} highlight the importance of representative and unbiased training sets in computer vision to avoid discrimination while \citet{bolukbasi} reveal that even the pre-trained word embeddings \textit{Word2Vec} contain societal gender bias. Model introspection goes hand in hand with other monitoring practices such as anomaly or drift detection. When an instance is flagged as an outlier, explanation methods can help determine whether the model prediction on the instance can be trusted and acted on.

The suitability of an explanation method depends on the data modality, type of model (e.g. tree-based or neural network) and prediction task (e.g. classification or regression). Most commonly used explanation methods also need heuristics and make assumptions about key components of the explanation generating process such as the background values for each feature (e.g. SHAP \cite{lundberg2017}, Integrated Gradients \cite{sundararajan2017}, Anchors \cite{ribeiro2018}, Contrastive Explanations \cite{dhurandhar2018}), local model behaviour (e.g. local linearity for LIME \cite{ribeiro2016}) or feature interactions (PDP \cite{pdp2001}). In order to overcome method-specific pitfalls, it is important to obtain a holistic explanation for each instance, combining the complementary strengths of different techniques. The overall explanation sheds light on the impact of the training data, relative feature importance, the key features to maintain the original prediction as well as the minimal changes to the features that will cause the prediction to change. Explanation techniques based on influence functions \cite{koh17, relatif} highlight which training instances had the most impact on a specific prediction at inference time. This allows the user to check whether the most influential training data contain relevant features compared to the explained instance in production. We also want to know which features are key to ensure a model prediction for a given instance regardless of the values of the other features in the form of Anchor explanations \cite{ribeiro2018}. The opposite and complementary counterfactual approach \cite{wachter2018} finds the minimal change to the original instance which flips the model prediction while still respecting the class-conditional data distribution \cite{looveren2019interpretable}. Finally, feature attribution methods such as LIME, SHAP or Integrated Gradients evaluate the relative feature importances with respect to a model prediction.

Explanation algorithms can be grouped into \textit{white-box} and \textit{black-box} methods. Whereas white-box methods assume access to model internals such as being able to take gradients with respect to the input \cite{sundararajan2017}, black-box methods do not assume anything beyond being able to access the prediction API endpoint. This is a natural scenario for deployed models as any information related to model internals is typically inscrutable and the only way to interact with the model is by requesting predictions. In the following section we motivate the usage of black-box explanations for models in production.

\subsection{Deploying black-box explanations}\label{sec:bb-exp}
\begin{figure}[htbp]
    \centering
    \includegraphics[width=\columnwidth]{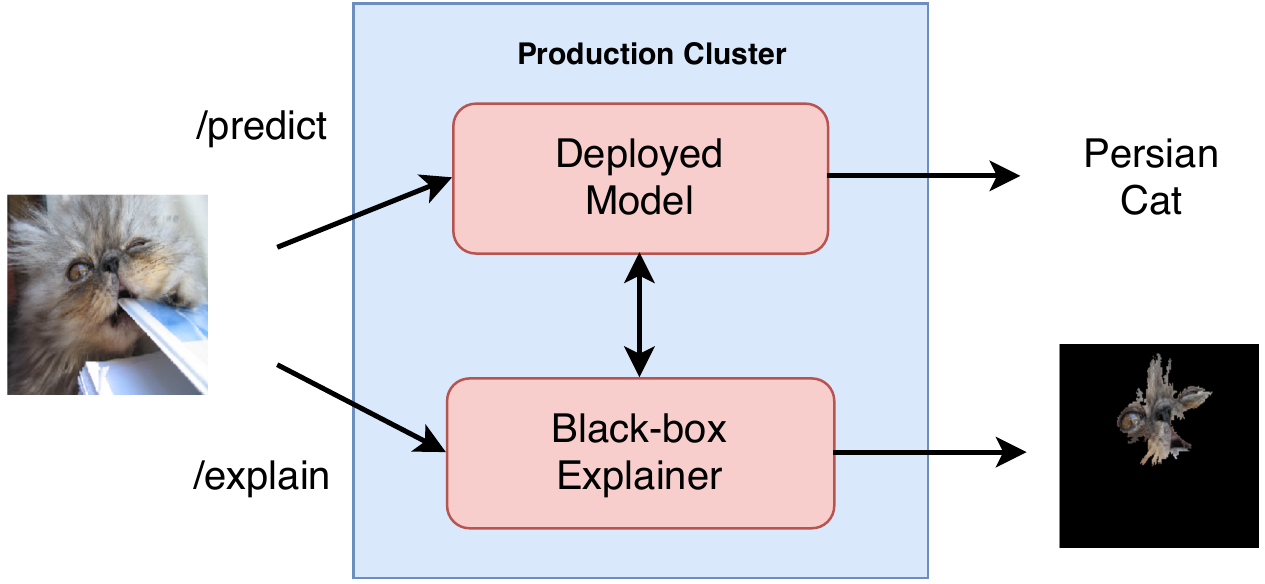}
    \caption{Design pattern for black-box explanation methods in production. A production cluster contains a deployed model which exposes a \texttt{/predict} endpoint as well as a deployed explainer which stores a reference to the model's \texttt{/predict} endpoint and exposes an additional \texttt{/explain} endpoint. An explanation can be requested on-demand by calling the \texttt{/explain} endpoint which triggers the black-box algorithm to interact with the model and produce an explanation.}
    \label{fig:exp_pattern}
\end{figure}

Black-box explanation algorithms work by taking an input instance whose prediction is to be explained and by repeatedly querying the model with modified versions of the input to approximate its predictive behaviour. The actual query strategy, definition of modified instances and explanation output is specific to each algorithm. In production this translates to having two deployments, the original model and the explainer, exposing a prediction endpoint and an explanation endpoint respectively (\cref{fig:exp_pattern}). When the prediction endpoint is called with a data point, a prediction is returned as usual, but when the explanation endpoint is called with the same data point, this triggers the black-box explanation algorithm to internally query the model and produce an explanation.

This pattern provides an infrastructure abstraction for requesting on-demand explanations for any black-box explanation algorithm. It has the advantage of using the underlying infrastructure to auto-scale the model deployment if a high volume of explanations is requested. Alternatively, this pattern can be implemented on a carbon copy of the model so as to separate production prediction requests from introspective explanation requests. This pattern is implemented using the Seldon Core \cite{seldon-core} and KFServing \cite{kfserving} deployment platforms, together with the Anchor explanation technique \cite{ribeiro2018} implemented in Alibi \cite{alibi} on tabular, text and image classification tasks. \Cref{fig:exp_pattern} shows an example explanation of a prediction made by a deployed InceptionV3 \cite{inceptionv3} model on an ImageNet \cite{imagenet_cvpr09} instance.

\section{Conclusion}\label{sec:conclusion}
In this paper we discussed key areas and challenges surrounding monitoring and explaining deployed models. We highlighted open source solutions for the algorithmic challenges \cite{alibi, alibi-detect, creme} and the infrastructure \cite{seldon-core, kfserving} to support these capabilities. One of the main open research topics is to more directly relate the label-independent measures obtained from the metrics, outlier and drift detectors to the model performance. Best practices are still being established in the MLOps community and we feel that in order to succeed, the open source tools need to be general enough to cover a majority of use cases whilst being flexible enough to allow for use case specific customization.


\bibliography{references}
\bibliographystyle{icml2020}

\end{document}